\documentstyle[colap,epsf]{article} 

\title{\vspace{-0.5in} Learning Dependencies between Case Frame Slots}
\author{Hang Li and Naoki Abe \\ Theory NEC Laboratory,
  RWCP\thanks{Real World Computing Partnership} \\ c/o C\&C Research Laboratories, NEC \\ 4-1-1 Miyazaki
  Miyamae-ku, Kawasaki, 216 Japan \\ 
  \{lihang,abe\}@sbl.cl.nec.co.jp } 

\newcommand{\tab}{\hspace{5mm}}

\begin{document}
\maketitle
\vspace{-0.5in}
\begin{abstract}
  We address the problem of automatically acquiring case frame
  patterns (selectional patterns) from large corpus data. In
  particular, we propose a method of learning dependencies between
  case frame slots. We view the problem of learning case frame
  patterns as that of learning multi-dimensional discrete joint
  distributions, where random variables represent case slots. We then
  formalize the dependencies between case slots as the {\em
    probabilistic} dependencies between these random variables. Since
  the number of parameters in a multi-dimensional joint distribution
  is exponential, it is infeasible to accurately estimate them in
  practice. To overcome this difficulty, we settle with approximating
  the target joint distribution by the product of {\em low} order
  component distributions, based on corpus data. In particular we
  propose to employ an efficient learning algorithm based on the MDL
  principle to realize this task. Our experimental results indicate
  that for certain classes of verbs, the accuracy achieved in a
  disambiguation experiment is improved by using the acquired
  knowledge of dependencies.
\end{abstract}

\section{Introduction}

We address the problem of automatically acquiring case frame patterns
(selectional patterns) from large corpus data. The acquisition of
case frame patterns normally involves the following three subproblems:
1) extracting case frames from corpus data, 2) generalizing case frame
slots within the case frames, 3) learning dependencies that exist
between the (generalized) case frame slots.

In this paper, we propose a method of learning dependencies between
case frame slots. By `dependency' is meant the relation that exists
between case slots which constrains the possible values assumed by
each of those case slots. As illustrative examples, consider the
following sentences.
\begin{equation}
\mbox{The girl will fly a jet.}
\end{equation}
\begin{equation}
\mbox{This airline company flies many jets.}
\end{equation}
\begin{equation}
\mbox{The girl will fly Japan Airlines.}
\end{equation}
\begin{equation}
\mbox{*The airline company will fly Japan Airlines.}
\end{equation}
We see that an `airline company' can be the subject of verb `fly' (the
value of slot `arg1'), when the direct object (the value of slot
`arg2') is an `airplane' but not when it is an `airline company.' 
These examples indicate that the possible values of case slots depend
in general on those of the other case slots: that is, there exist
`dependencies' between different case slots.

The knowledge of such dependencies is useful in various tasks in
natural language processing, especially in analysis of sentences
involving multiple prepositional phrases, such as
\begin{equation}\label{hold}
\begin{array}{l}
  \mbox{The girl will fly a jet from Tokyo to Beijing.}
\end{array}
\end{equation}
Note in the above example that the slot of `from' and that of `to'
should be considered dependent and the attachment site of one of the
prepositional phrases (case slots) can be determined by that of the
other with high accuracy and confidence.

There has been no method proposed to date, however, that learns
dependencies between case slots in the natural language processing
literature. In the past research, the distributional pattern of each
case slot is learned independently, and methods of resolving
ambiguities are also based on the assumption that case slots are
independent\cite{Hindle91,Chang92,Sekine92,Resnik93a,Grishman94,Alshawi95,Li95},
or at most two case slots are dependent
\cite{Brill94,Ratnaparkhi94,Collins95}. Thus, provision of an
effective method of learning dependencies between case slots, as well
as investigation of the usefulness of the acquired dependencies in
disambiguation and other natural language processing tasks would be an
important contribution to the field.

In this paper, we view the problem of learning case frame patterns as
that of learning multi-dimensional discrete joint distributions, where
random variables represent case slots. We then formalize the
dependencies between case slots as the {\em probabilistic}
dependencies between these random variables. Since the number of
parameters that exist in a multi-dimensional joint distribution is
exponential if we allow n-ary dependencies, it is infeasible to
accurately estimate them with a data size available in practice. It is
also clear that relatively few of these random variables (case slots)
are actually dependent on each other with any significance. Thus it is
likely that the target joint distribution can be approximated
reasonably well by the product of component distributions of {\em low}
order, drastically reducing the number of parameters that need to be
considered. This is indeed the approach we take in this paper.

Now the problem is how to approximate a joint distribution by the
product of low order component distributions. Recently, Suzuki
proposed an algorithm to approximately learn a multi-dimensional
discrete joint distribution expressible as a `dendroid distribution,'
which is both efficient and theoretically sound \cite{Suzuki93}. We
employ Suzuki's algorithm to learn case frame patterns as dendroid
distributions. We conducted some experiments to automatically acquire
case frame patterns from the Penn Tree Bank bracketed corpus. Our
experimental results indicate that for some classes of verbs the
accuracy achieved in a disambiguation experiment can be improved by
using the acquired knowledge of dependencies between case slots.

\section{Probabilistic Models for Case Frame Patterns}

\begin{figure}[htp]
\begin{center}
\begin{verbatim}
(fly (arg1 girl)(arg2 jet))
(fly (arg1 company)(arg2 jet))
(fly (arg1 girl)(arg2 company))
\end{verbatim}
\caption{Example case frames generated by a word-based model}
\label{fig:word}
\end{center}
\end{figure}

Suppose that we have data of the type shown in Figure \ref{fig:word},
given by case frame instances of verb `fly' automatically extracted
from a corpus, using conventional techniques. As explained in
Introduction, the problem of learning case frame patterns can be
viewed as that of estimating the underlying multi-dimensional joint
discrete distributions which give rise to such data. In this research,
we assume that case frame instances with the same head are generated
by a joint distribution of type,
\begin{equation}
P_{Y}(X_1,X_2,\ldots,X_n),
\end{equation}
where index $Y$ stands for the head, and each of the random variables
$X_i, i=1,2,\ldots,n,$ represents a case slot. In this paper, we use
`case slots' to mean {\em surface} case slots, and we uniformly treat
obligatory cases and optional cases. Thus the number $n$ of the
random variables is roughly equal to the number of prepositions in
English (and less than $100$).

\begin{figure*}[htb]
\begin{verbatim}
<...> : word class
(fly (arg1 <person>)(arg2 <airplane>))
(fly (arg1 <person>)(arg2 <airplane>))
(fly (arg1 <person>)(arg2 <airplane>))
(fly (arg1 <company>)(arg2 <airplane>))
(fly (arg1 <company>)(arg2 <airplane>))
(fly (arg1 <person>)(arg2 <company>))
(fly (arg1 <person>)(to <place>))
(fly (arg1 <person>)(from <place>)(to <place>))
(fly (arg1 <company>)(from <place>)(to <place>))
\end{verbatim}
\caption{Example case frames generated by a class-based model}
\label{fig:class}
\end{figure*}

These models can be further classified into three types of
probabilistic models according to the type of values each random
variable $X_i$ assumes. When $X_i$ assumes a word or a special symbol
`$0$' as its value, we refer to the corresponding model
$P_Y(X_1,\ldots,X_n)$ as a `word-based model.'  Here `$0$' indicates
the absence of the case slot in question. When $X_i$ assumes a
word-class or `$0$' as its value, the corresponding model is called a
`class-based model.' When $X_i$ takes on $1$ or $0$ as its value, we
call the model a `slot-based model.'  Here the value of `$1$'
indicates the presence of the case slot in question, and `$0$'
absence. For example, the data in Figure \ref{fig:word} can be
generated by a word-based model, and the data in Figure
\ref{fig:class} by a class-based model. Suppose for simplicity that
there are only 4 possible case slots corresponding respectively to the
subject, direct object, `from' phrase, and `to' phrase. Then,
\begin{equation}
\begin{array}{l}
  P_{fly}(X_{arg1}={\rm girl},X_{arg2}={\rm jet}, X_{from}=0,X_{to}=0)
\end{array}
\end{equation}
is given a specific probability value by a word-based model. In
contrast,
\begin{equation}
\begin{array}{l}
  P_{fly}(X_{arg1}=\langle{\rm person}\rangle,X_{arg2}=\langle{\rm
    airplane}\\
 \rangle,X_{from}=0,X_{to}=0)
\end{array}
\end{equation}
is given a specific probability value by a class-based model, where
$\langle{\rm person}\rangle$ and $\langle{\rm airplane}\rangle$ denote
word classes. Finally,
\begin{equation}
\begin{array}{l}
  P_{fly}(X_{arg1}=1,X_{arg2}=1,X_{from}=0,X_{to}=0)
\end{array}
\end{equation}
is assigned a specific probability value by a slot-based model.  We
then formulate the dependencies between case slots as the
probabilistic dependencies between the random variables in each of
these three models.

In the absence of any constraints, however, the number of parameters
in each of the above three models is exponential (even the slot-based
model has $O(2^{n})$ parameters ), and thus it is infeasible to
accurately estimate them in practice. A assumption that is often made
to deal with this difficulty is that random variables (case slots) are
mutually independent.

Suppose for example that in the analysis of the sentence
\begin{equation}
\mbox{The girl will fly a jet from Tokyo,}
\end{equation}
the following alternative interpretations are given.
\begin{equation}\label{eq:int1}
\mbox{(fly (arg1 {\rm girl}) (arg2 {\rm jet}) (from {\rm Tokyo}))}
\end{equation}
\begin{equation}\label{eq:int2}
\mbox{(fly (arg1 {\rm girl}) (arg2 ({\rm jet} (from {\rm Tokyo}))))}
\end{equation}
We wish to select the more appropriate of the two interpretations. A
{\em heuristic} word-based method for disambiguation, in which the
random variables (case slots) are assumed to be dependent, is to
calculate the following values of word-based likelihood and to select
the interpretation with the higher likelihood value.
\begin{equation}
P_{fly}(X_{arg1}={\rm girl},X_{arg2}={\rm jet},X_{from}={\rm Tokyo})
\end{equation}
\begin{equation}
\begin{array}{l}
P_{fly}(X_{arg1}={\rm girl},X_{arg2}={\rm jet}) \\
\times P_{jet}(X_{from}={\rm Tokyo})
\end{array}
\end{equation}
If on the other hand we assume that the random variables are {\em
  independent}, we only need to calculate and compare
(c.f.\cite{Li95})
\begin{equation}
P_{fly}(X_{from}={\rm Tokyo})
\end{equation}
and
\begin{equation}
P_{jet}(X_{from}={\rm Tokyo}).
\end{equation}
The independence assumption can also be made in the case of a
class-based model or a slot-based model. For slot-based models, with
the independence assumption, the following probabilities
\begin{equation}
P_{fly}(X_{from} = 1)
\end{equation}
\begin{equation}
P_{jet}(X_{from} = 1)
\end{equation}
are to be compared (c.f.\cite{Hindle91}). 

Assuming that random variables (case slots) are mutually independent
would drastically reduce the number of parameters. (Note that under
the independence assumption the number of parameters in a slot-based
model becomes $O(n)$.)  As illustrated in Section 1, this assumption
is not necessarily valid in practice. 

What seems to be true in practice is that some case slots are in fact
dependent but overwhelming majority of them are independent, due
partly to the fact that usually only a few case slots are obligatory
and most others are optional.\footnote{Optional case slots are not
  necessarily independent, but if two optional case slots are randomly
  selected, it is likely that they are independent of one another.}
Thus the target joint distribution is likely to be approximable by the
product of several component distributions of low order, and thus have
in fact a reasonably small number of parameters. We are thus lead to
the approach of approximating the target joint distribution by such a
simplified model, based on corpus data.

\section{Approximation by Dendroid Distribution}

Without loss of generality, any n-dimensional joint distribution can
be written as
\begin{equation}
P(X_1,X_2,\ldots,X_n) = \prod_{i=1}^{n} P(X_{m_i}|X_{m_1}...X_{m_{i-1}})
\end{equation}
for some permutation ($m_1,m_2,...m_n$) of $1,2,..,n$, where we let
$P(X_{m_1}|X_{m_0})$ denote $P(X_{m_1})$.

A plausible assumption on the dependencies between random variables is
intuitively that each variable {\em directly} depends on at most one
other variable. (Note that this assumption is the simplest among those
that relax the independence assumption.) For example, if a joint
distribution $P(X_1,X_2,X_3)$ over 3 random variables $X_1,X_2,X_3$
can be written (approximated) as follows, it (approximately) satisfies
such an assumption.
\begin{equation}
  P(X_1,X_2,X_3) =(\approx) P(X_1)\times P(X_2|X_1)\times P(X_3|X_1)
\end{equation}
Such distributions are referred to as `dendroid distributions' in the
literature. A dendroid distribution can be represented by a dependency
forest (i.e. a set of dependency trees), whose nodes represent the
random variables, and whose directed arcs represent the dependencies
that exist between these random variables, each labeled with a number
of parameters specifying the probabilistic dependency. (A dendroid
distribution is a restricted form of the Bayesian network
\cite{Pearl88}.)  It is not difficult to see that there are 7 and only
7 such representations for the joint distribution $P(X_1,X_2,X_3)$
(See Figure \ref{fig:graph}), disregarding the actual numerical values
of the probabilistic parameters.

Now we turn to the problem of how to select the best dendroid
distribution from among all possible ones to approximate a target
joint distribution based on input data `generated' by it. This problem
has been investigated in the area of machine learning and related
fields.  A classical method is Chow \& Liu's algorithm for estimating
a multi-dimensional joint distribution as a dependency tree, in a way
which is both efficient and theoretically sound \cite{Chow68}. More
recently Suzuki extended their algorithm so that it estimates the
target joint distribution as a dendroid distribution or dependency
forest \cite{Suzuki93}, allowing for the possibility of learning one
group of random variables to be completely independent of another.
Since many of the random variables (case slots) in a case frame
pattern are essentially independent, this feature is crucial in our
context, and we thus employ Suzuki's algorithm for learning our case
frame patterns.

\setlength{\unitlength}{1mm}
\begin{figure*}[htb]
\begin{center}
\begin{picture}(150,120)(0,0)
\put(20,40){$X_1$}
\put(10,20){$X_2$}
\put(30,20){$X_3$}
\put(5,15){$P(X_1)P(X_2\vert X_1)P(X_3\vert X_2)$}
\put(1,10){$=P(X_2)P(X_1\vert X_2)P(X_3\vert X_2)$}
\put(1,5){$=P(X_3)P(X_2\vert X_3)P(X_1\vert X_2)$}
\put(15,21){\line(1,0){15}}
\put(22,39){\line(-1,-2){8}}

\put(70,40){$X_1$}
\put(60,20){$X_2$}
\put(80,20){$X_3$}
\put(55,15){$P(X_2)P(X_1\vert X_2)P(X_3\vert X_1)$}
\put(51,10){$=P(X_1)P(X_3\vert X_1)P(X_2\vert X_1)$}
\put(51,5){$=P(X_3)P(X_1\vert X_3)P(X_2\vert X_1)$}
\put(72,39){\line(-1,-2){8}}
\put(72,39){\line(1,-2){8}}

\put(120,40){$X_1$}
\put(110,20){$X_2$}
\put(130,20){$X_3$}
\put(105,15){$P(X_1)P(X_3\vert X_1)P(X_2\vert X_3)$}
\put(101,10){$=P(X_3)P(X_1\vert X_3)P(X_2\vert X_3)$}
\put(101,5){$=P(X_2)P(X_3\vert X_2)P(X_1\vert X_3)$}
\put(130,21){\line(-1,0){15}}
\put(122,39){\line(1,-2){8}}

\put(20,85){$X_1$}
\put(10,65){$X_2$}
\put(30,65){$X_3$}
\put(5,60){$P(X_1)P(X_2)P(X_3\vert X_2)$}
\put(1,55){$=P(X_1)P(X_3)P(X_2\vert X_3)$}
\put(15,66){\line(1,0){15}}

\put(70,85){$X_1$}
\put(60,65){$X_2$}
\put(80,65){$X_3$}
\put(55,60){$P(X_1)P(X_2\vert X_1)P(X_3)$}
\put(51,55){$=P(X_2)P(X_1\vert X_2)P(X_3)$}
\put(72,84){\line(-1,-2){8}}

\put(120,85){$X_1$}
\put(110,65){$X_2$}
\put(130,65){$X_3$}
\put(105,60){$P(X_1)P(X_3\vert X_1)P(X_2)$}
\put(101,55){$=P(X_3)P(X_1\vert X_3)P(X_2)$}
\put(122,84){\line(1,-2){8}}

\put(20,125){$X_1$}
\put(10,105){$X_2$}
\put(30,105){$X_3$}
\put(5,100){$P(X_1)P(X_2)P(X_3)$}
\end{picture}
\caption{Dendroid distributions}
\label{fig:graph}
\end{center}
\end{figure*}

Suzuki's algorithm first calculates the mutual information between all
two nodes (random variables), and it sorts the node pairs in
descending order with respect to the mutual information. It then puts
a link between a node pair with the largest mutual information value
$I$, provided that $I$ exceeds a certain threshold which depends on
the node pair and adding that link will not create a loop in the
current dependency graph. It repeats this process until no node pair
is left unprocessed. Figure \ref{fig:algorithm} shows the detail of
this algorithm, where $k_i$ denotes the number of possible values
assumed by $X_i$, $N$ the input data size, and $\log$ denotes the
logarithm to the base $2$.
\begin{figure*}[htb]
\begin{tabbing}
{\bf Algorithm:} \\
1. Let $T:= \emptyset$; \\
2. Calculate the mutual information $I(X_i,X_j)$ for all node pairs $(X_i,X_j)$
; \\
3. Sort the node pairs in descending order of $I$, and store them into queue $Q
$; \\
4. Let $V$ be the set of $\{ X_i\}$, $i=1,2,...,n$; \\
5. {\bf while} the maximum value of $I$ of the node pair $(X_i,X_j)$
in $Q$ satisfies \\
\tab $I(X_i,X_j) > \theta(X_i,X_j) = (k_i-1) (k_j-1) \frac{\log N}{2\cdot N}$ \\
\tab {\bf do} \\
\tab {\bf begin} \\
\tab 4-1. Remove the node pair $(X_i,X_j)$ having the maximum value of $I$ from $Q$; \\
\tab 4-2. {\bf if} $X_i$ and $X_j$ belong to different sets
$W_1$,$W_2$ in $V$; \\
\tab \tab {\bf then} \\
\tab \tab Replace $W_1$ and $W_2$ in $V$ with $W_1 \cup W_2$, and add
edge $(X_i,X_j)$ to $T$; \\
\tab {\bf end} \\
6. Output $T$ as the set of edges of the estimated model. \\
\end{tabbing}
\caption{The learning algorithm}
\label{fig:algorithm}
\end{figure*}
It is easy to see that the number of parameters in a dendroid
distribution is of the order $O(n k^2)$, where $k$ is the maximum of
all $k_i$, and $n$ is the number of random variables. The time
complexity of the algorithm is of the order $O(n^2(k^2+\log n))$.

We will now show how the algorithm works by an illustrative example.
Suppose that the data is given as in Figure \ref{fig:class} and there
are 4 nodes (random variables) $X_{arg1},X_{arg2},X_{from},X_{to}$.
The values of mutual information and thresholds for all node pairs are
shown in Table \ref{tb:mutual}.\footnote{The probabilities in this
  table are estimated by using the so-called Expected Likelihood
  Estimator, i.e., by adding $0.5$ to actual frequencies (c.f.
  \cite{Gale90}).} Based on this calculation the algorithm constructs
the dependency forest shown in Figure~\ref{fig:example}, because the
mutual information between $X_{arg2}$ and $X_{to}$ , $X_{from}$ and
$X_{to}$ are large enough, but not the others. The result indicates
that slot `arg2' and `from' should be considered dependent on `to.'
Note that `arg2' and `from' should also be considered dependent via
`to' but to a somewhat weaker degree.
\begin{table*}[htb]
\begin{center}
\caption{Mutual information and threshold values for node pairs}
\label{tb:mutual}
\begin{tabular}{|l|l|l|l|l|} \hline
$I:\theta$ & $X_{arg1}$ & $X_{arg2}$ & $X_{from}$ & $X_{to}$ \\ \hline
$X_{arg1}$ & & $0.01:0.35$ & $0.02:0.18$ & $0.00:0.18$ \\
$X_{arg2}$ & & $$ & $0.22:0.35$ & \underline{$0.43:0.35$} \\
$X_{from}$ & & $$ & $$ & \underline{$0.26:0.18$} \\
$X_{to}$ & & $$ & $$ & $$ \\ \hline
\end{tabular}
\end{center}
\end{table*}
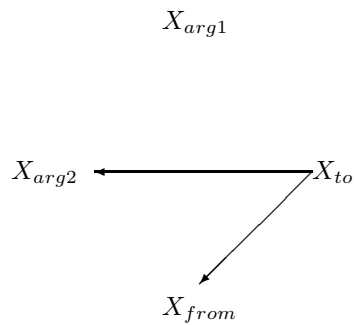
\begin{figure}[htb]
\begin{center}
\begin{picture}(80,60)(0,0)
\put(30,50){$X_{arg1}$}
\put(10,30){$X_{arg2}$}
\put(30,12){$X_{from}$}
\put(50,30){$X_{to}$}
\put(50,31){\vector(-1,0){29}}
\put(50,31){\vector(-1,-1){15}}
\end{picture}
\caption{An example case frame pattern}
\label{fig:example}
\end{center}
\end{figure}

Suzuki's algorithm is derived from the Minimum Description Length
(MDL) principle
\cite{Rissanen78,Rissanen83,Rissanen84,Rissanen86,Rissanen89} which is
a principle for data compression and estimation from information
theory and statistics. It is known that as a strategy of estimation,
MDL is guaranteed to be near optimal.\footnote{We refer the interested
  reader to \cite{Quinlan89,Li95} for an introduction to MDL.} In
applying MDL, we usually assume that the given data are generated by a
probabilistic model that belongs to a certain class of models and
selects a model within the class which best explains the data. It
tends to be the case usually that a simpler model has a poorer fit to
the data, and a more complex model has a better fit to the data. Thus
there is a trade-off between the simplicity of a model and the
goodness of fit to data. MDL resolves this trade-off in a disciplined
way: It selects a model which is reasonably simple and fits the data
satisfactorily as well. In our current problem, a simple model means a
model with less dependencies, and thus MDL provides a theoretically
sound way to learn only those dependencies that are statistically
significant in the given data. An especially interesting feature of
MDL is that it incorporates the input data size in its model selection
criterion. This is reflected, in our case, in the derivation of the
threshold $\theta$. Note that when we do not have enough data (i.e. 
for small $N$), the thresholds will be large and few nodes tend to be
linked, resulting in a simple model in which most of the case slots
are judged independent.  This is reasonable since with a small data
size most case slots cannot be determined to be dependent with any
significance.

\section{Experimental Results}

We conducted some experiments to test the performance of the proposed
method as a method of acquiring case frame patterns. In particular, we
tested to see how effective the patterns acquired by our method are in
a structural disambiguation experiment. We will describe the results
of this experimentation in this section.

\subsection{Experiment 1: Slot-based Model}

In our first experiment, we tried to acquire case frame patterns as
slot-based models.  We extracted 181,250 case frames from the Wall
Street Journal (WSJ) bracketed corpus of the Penn Tree Bank
\cite{Marcus93} as training data. There were $357$ verbs for which
more than $50$ case frame examples appeared in the training data.
Table~\ref{tb:verb} shows the verbs that appeared in the data most
frequently and the numbers of their occurrences.

\begin{table}
\begin{center}
\caption{Verbs appearing most frequently}
\label{tb:verb}
\begin{tabular}{|lc|} \hline
Verb & Num. of frames \\ \hline
be & 17713 \\
say & 9840 \\
have & 4030 \\
make & 1770 \\
take & 1245 \\
expect & 1201 \\
sell & 1147 \\
rise & 1125 \\
get & 1070 \\
go & 1042 \\
do & 982 \\
buy & 965 \\
fall & 862 \\
add & 740 \\
come & 733 \\
include & 707 \\
give & 703 \\
pay & 700 \\
see & 680 \\
report & 674 \\ \hline
\end{tabular}
\end{center}
\end{table}

First we acquired the case frame patterns as slot-based models for all
of the $357$ verbs. We then conducted a ten-fold cross validation to
evaluate the `test data perplexities'\footnote{The `test data
  perplexity,' which is a measure of testing how well an estimated
  probabilistic model predicts some hitherto unseen data, is defined
  as $2^{H(P_T,P_M)}, H(P_T,P_M) = - \sum_{x} P_T(x) \cdot \log
  P_M(x)$, where $P_M(x)$ denotes the probability function of the
  estimated model, $P_T(x)$ the distribution function of the data
  \cite{Bahl83}.} of the acquired case frame patterns, that is, we
used nine tenth of the case frames for each verb as training data
(saving what remains as test data), to acquire case frame pattern for
the verb, and then calculated perplexity using the test data. We
repeated this process ten times and calculated the average perplexity.
Table \ref{tb:per} shows the average perplexities obtained for some
randomly selected verbs. We also calculated the average perplexities of
the `independent slot models' acquired based on the assumption that
each case slot is independent.  Our experimental results shown in
Table~\ref{tb:per} indicate that the use of the dendroid models can
achieve up to $20\%$ perplexity reduction as compared to the use of
the independent slot models. It seems safe to say therefore that the
dendroid model is more suitable for representing the {\em true} model
of case frames than the independent slot model.

\begin{table*}
\begin{center}
\caption{Verbs and their perplexities}
\label{tb:per}
\begin{tabular}{|lcc|} \hline
Verb & Independent & Dendroid(Reduction in percentage) \\ \hline
add & $5.82$ & $5.36(8\%)$ \\
buy & $5.04$ & $4.98(1\%)$ \\
find & $2.07$ & $1.92(7\%)$ \\
open & $20.56$ & $16.53(20\%)$ \\
protect & $3.39$ & $3.13(8\%)$ \\
provide & $4.46$ & $4.13(7\%)$ \\
represent & $1.26$ & $1.26(0\%)$ \\
send & $3.20$ & $3.29(-3\%)$ \\
succeed & $2.97$ & $2.57(13\%)$ \\
tell & $1.36$ & $1.36(0\%)$ \\ \hline
\end{tabular}
\end{center}
\end{table*}

We also used the acquired dependency knowledge in a pp-attachment
disambiguation experiment. We used the case frames of all $357$ verbs
as our training data. We used the entire bracketed corpus as training
data because we wanted to utilize as many training data as possible.
We extracted ($verb$,$noun_1$,$prep$,$noun_2$) and
($verb$,$prep_1$,$noun_1$,$prep_2$,$noun_2$) patterns from the WSJ
tagged corpus as test data, using pattern matching techniques such as
that described in \cite{Smadja93}. We took care to ensure that only
the part of the tagged (non-bracketed) corpus which does not overlap
with the bracketed corpus is used as test data. (The bracketed corpus
does overlap with part of the tagged corpus.)

We acquired case frame patterns using the training data. Figure
\ref{fig:buy} shows an example of the results, which is part of the
case frame pattern (dendroid distribution) for the verb `buy.'  Note
in the model that the slots `for,' 'on,' etc, are dependent on `arg2,'
while `arg1' and `from' are independent.

\begin{figure*}[htb]
{\small
\begin{verbatim}
buy:
[arg1]: [P(arg1=0)=0.000571] [P(arg1=1)=0.999429]
[arg2]: [P(arg2=0)=0.055114] [P(arg2=1)=0.944886]
           [P(on=1|arg2=1)= 0.018630] [P(on=0|arg2=1)= 0.981370]
           [P(on=1|arg2=0)= 0.112245] [P(on=0|arg2=0)= 0.887755]
           [P(for=1|arg2=1)= 0.109976] [P(for=0|arg2=1)= 0.890024]
           [P(for=1|arg2=0)= 0.255102] [P(for=0|arg2=0)= 0.744898]
           [P(by=1|arg2=1)= 0.004207] [P(by=0|arg2=1)= 0.995793]
           [P(by=1|arg2=0)= 0.051020] [P(by=0|arg2=0)= 0.948980]
[on]: [P(on=0)=0.976705] [p(on=1)=0.023295]
[for]: [P(for=0)=0.882386] [P(for=1)=0.117614]
[by]: [P(by=0)=0.993750] [P(by=1)=0.006250]
[from]: [P(from=0)=0.933523] [P(from=1)=0.066477]
\end{verbatim}
}
\caption{An example case frame pattern (dendroid distribution)}
\label{fig:buy}
\end{figure*}

We found that there were $266$ verbs, whose `arg2' slot is dependent
on some of the other preposition slots. Table~\ref{tb:depen1} shows
$37$ of the verbs whose dependencies between arg2 and other case slots
are positive and exceed a certain threshold, i.e. $P(arg2=1,prep=1) >
0.25$.\footnote{We uniformly treat the head of a noun phrase
  immediately after a verb as `arg2' (including, for example `30\%' in
  `rise 30\% from billion').} The dependencies found by our method
seem to agree with human intuition in most cases.

\begin{table*}
\begin{center}
\caption{Verbs and their dependent case slots}
\label{tb:depen1}
\begin{tabular}{|l|l|l|}\hline
Verb & Dependent slots & Example \\ \hline
achieve & arg2 \ in & achieve breakthrough in 1987 \\
acquire & arg2 \ in & acquire share in market \\
add & arg2 \ to & add 1 to 3 \\
begin & arg2 \ in & begin proceeding in London \\
blame & arg2 \ for & blame school for limitation \\
buy & arg2 \ for & buy property for cash \\
charge & arg2 \ in & charge man in court \\
climb & arg2 \ from & climb 20\% from million \\
compare & arg2 \ with & compare profit with estimate \\
convert & arg2 \ to & convert share to cash \\
defend & arg2 \ against & defend themselves  against takeover \\
earn & arg2 \ on & earn billion on revenue \\
end & arg2 \ at & end day at 778 \\
explain & arg2 \ to & explain it to colleague \\
fall & arg2 \ in & fall 45\% in 1977\\
file & arg2 \ against & file suit against company \\
file & arg2 \ with & file issue with commission \\
finish & arg2 \ at & finish point at 22 \\
focus & arg2 \ on & focus attention on value \\
give & arg2 \ to & give business to firm \\
increase & arg2 \ to & increase number to five \\
invest & arg2 \ in & invest share in fund \\
negotiate & arg2 \ with & negotiate rate with advertiser \\
open & arg2 \ in & open bureau in capital \\
pay & arg2 \ for & pay million for service \\
play & arg2 \ in & play role in takeover \\
prepare & arg2 \ for & prepare case for trial \\
provide & arg2 \ for & provide engine for plane \\
pull & arg2 \ from & pull money from market \\
refer & arg2 \ to & refer inquiry to official \\
return & arg2 \ to & return car to dealer \\
rise & arg2 \ from & rise 10\% from billion \\
spend & arg2 \ on & spend money on production \\
surge & arg2 \ in & surge 10\% in 1988 \\
surge & arg2 \ to & surge 25\% to million \\
trade & arg2 \ in & trade stock in transaction \\
turn & arg2 \ to & turn ball to him \\
withdraw & arg2 \ from & withdraw application from office \\ \hline
\end{tabular}
\end{center}
\end{table*}

There were $93$ examples in the test data
($verb$,$noun_1$,$prep$,$noun_2$ pattern) in which the two slots
`arg2' and $prep$ of $verb$ are determined to be positively dependent
and their dependencies are stronger than the threshold of $0.25$. We
forcibly attached $prep~noun_2$ to $verb$ for these $93$ examples. For
comparison, we also tested the disambiguation method based on the
independence assumption proposed by \cite{Li95} on these examples.
Table~\ref{tb:result1} shows the results of these experiments, where
`Dendroid' stands for the former method and `Independent' the latter.
We see that using the information on dependency we can {\em
  significantly} improve the disambiguation accuracy on this part of
the data. Since we can use existing methods to perform disambiguation
for the rest of the data, we can improve the disambiguation accuracy
for the entire test data using this knowledge.

\begin{table}
\begin{center}
\caption{Disambiguation results 1}
\label{tb:result1}
\begin{tabular}{|l|c|} \hline
 & Accuracy($\%$) \\ \hline
Dendroid & $90/93(96.8)$ \\
Independent & $79/93(84.9)$ \\ \hline
\end{tabular}
\end{center}
\end{table}

Furthermore, we found that there were $140$ verbs having
inter-dependent preposition slots. Table~\ref{tb:depen2} shows $22$
out of these $140$ verbs such that their case slots have positive
dependency that exceeds a certain threshold, i.e.
$P(prep_1=1,prep_2=1) > 0.25 $. Again the dependencies found by our
method seem to agree with human intuition.

\begin{table*}[h]
\begin{center}
\caption{Verbs and their dependent case slots}
\label{tb:depen2}
\begin{tabular}{|l|l|l|} \hline
Head & Dependent slots & Example \\ \hline
acquire & from \ for & acquire from corp. for million \\
apply & for \ to & apply to commission for permission \\
boost & from \ to & boost from 1\% to 2\% \\
climb & from \ to & climb from million to million \\
climb & in \ to & climb to million in segment \\
cut & from \ to & cut from 700 to 200 \\
decline & from \ to & decline from billion to billion \\
end & at \ on & end at 95 on screen \\
fall & from \ to & fall from million to million \\
grow & from \ to & grow from million to million \\
improve & from \ to & improve from 10\% to 50\% \\
increase & from \ to & increase from million to million \\
jump & from \ to & jump from yen to yen \\
move & from \ to & move from New York to Atlanta \\
open & at \ for & open for trading at yen \\
raise & from \ to & raise from to 5\% to 10\% \\
reduce & from \ to & reduce from 5\% to 1\% \\
rise & from \ to & rise from billion to billion \\
sell & to \ for & sell to bakery for amount \\
shift & from \ to & shift from stock to bond \\
soar & from \ to & soar from 10\% to 15\% \\
think & of \ as & think of this as thing \\ \hline
\end{tabular}
\end{center}
\end{table*}

In the test data ($verb$,$prep_1$,$noun_1$,$prep_2$,$noun_2$ pattern),
there were $21$ examples that involves one of the above $22$ verbs
whose preposition slots show dependency exceeding $0.25$. We forcibly
attached both $prep_1~noun_1$ and $prep_2~noun_2$ to $verb$ on these
$21$ examples, since the two slots $prep_1$ and $prep_2$ are judged
dependent. Table~\ref{tb:result2} shows the results of this
experimentation, where `Dendroid' and `Independent' respectively
represent the method of using and not using the dependencies. Again,
we find that for the part of the test data in which dependency is
present, the use of the dependency knowledge can be used to improve
the accuracy of a disambiguation method, although our experimental
results are inconclusive at this stage.

\begin{table}
\begin{center}
\caption{Disambiguation results 2}
\label{tb:result2}
\begin{tabular}{|l|c|} \hline
 & Accuracy($\%$) \\ \hline
Dendroid & $21/21(100)$ \\
Independent & $20/21(95.2)$ \\ \hline
\end{tabular}
\end{center}
\end{table}

\subsection{Experiment 2: Class-based Model}

We also used the $357$ verbs and their case frames used in Experiment
1 to acquire case frame patterns as class-based models using the
proposed method. We randomly selected $100$ verbs among these $357$
verbs and attempted to acquire their case frame patterns. We
generalized the case slots within each of these case frames using the
method proposed by \cite{Li95} to obtain class-based case slots, and
then replaced the word-based case slots in the data with the obtained
class-based case slots.  What resulted are class-based case frame
examples like those shown in Figure~\ref{fig:class}. We used these
data as input to the learning algorithm and acquired a case frame
pattern for each of the $100$ verbs. We found that no two case slots
are determined as dependent in any of the case frame patterns. This is
because the number of parameters in a class based model is very large
compared to the size of the data we had available.

Our experimental result verifies the validity in practice of the
assumption widely made in statistical natural language processing that
class-based case slots (and also word-based case slots) are mutually
independent, at least when the data size available is that provided by
the current version of the Penn Tree Bank. This is an empirical
finding that is worth noting, since up to now the independence
assumption was based solely on human intuition, to the best of our
knowledge.

\begin{figure*}[htb]
\vspace{-0.5cm}
\begin{center}
\begin{tabular}{lll}
\hspace{-1cm}
&
{\epsfxsize8cm\epsfysize5cm\epsfbox{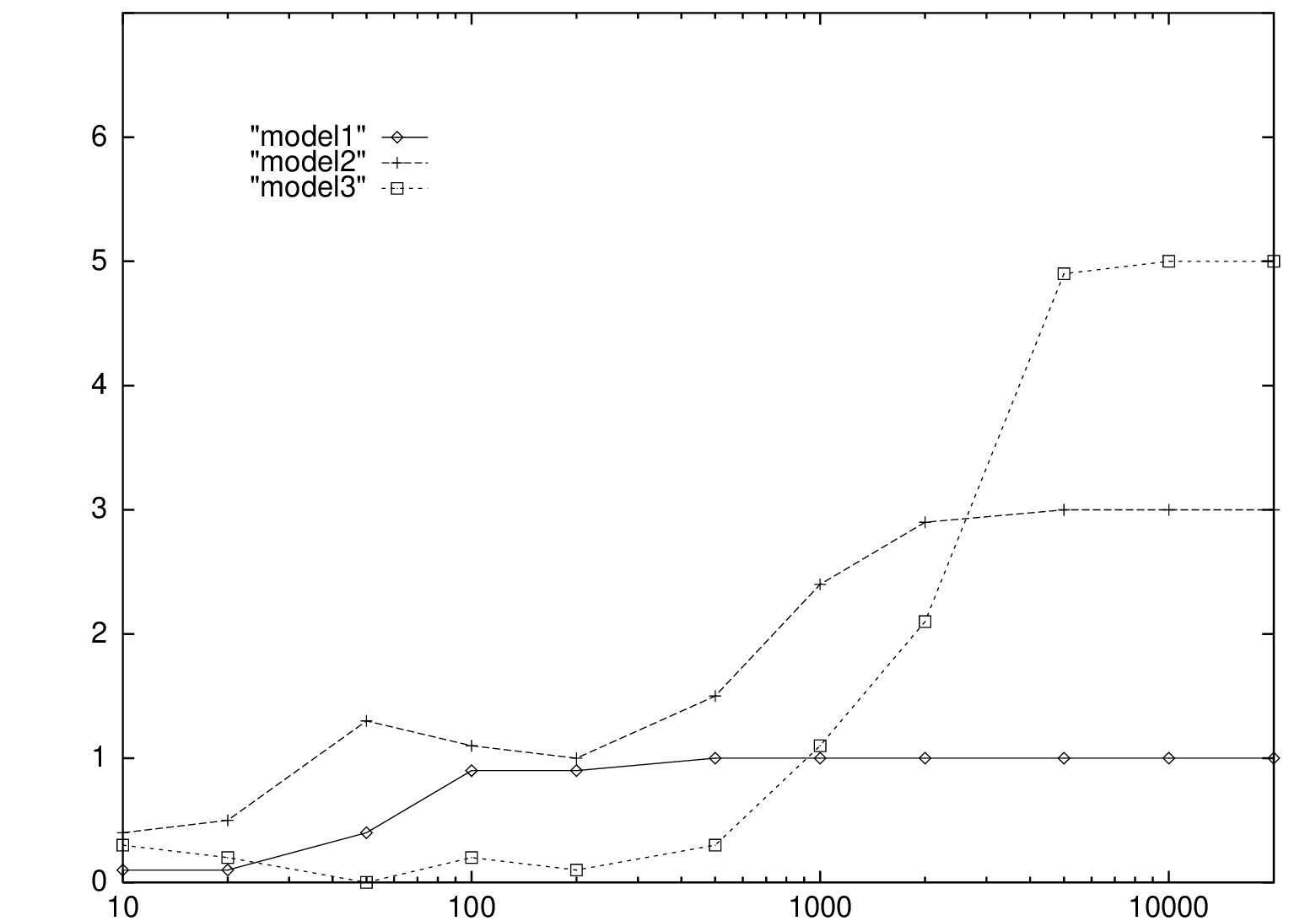}}
&
{\epsfxsize8cm\epsfysize5cm\epsfbox{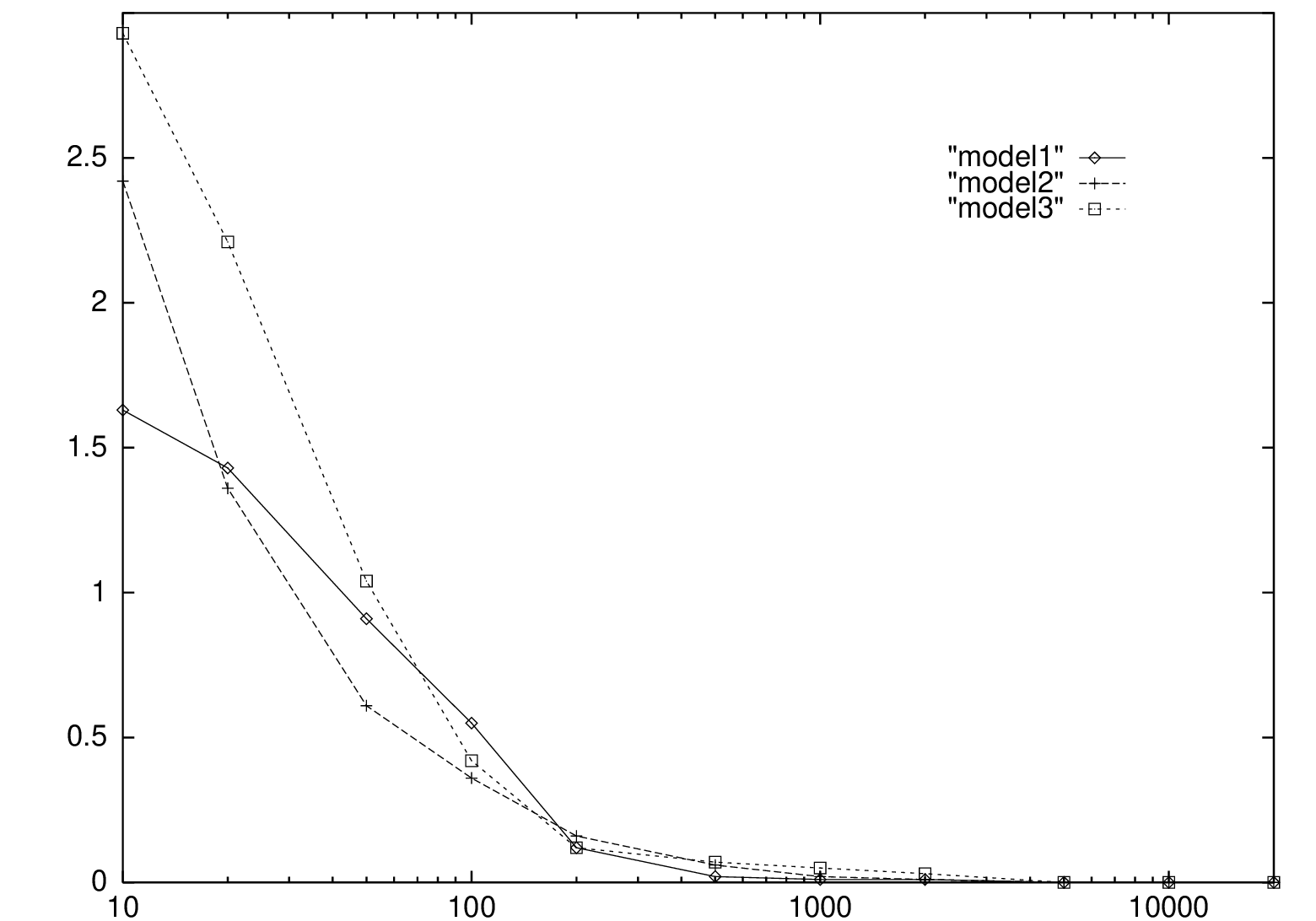}}
\end{tabular}
\vspace{-0.5cm} 
\caption{(a) Number of dependencies versus data size and
(b) KL distance versus data size}
\label{fig:sim} 
\end{center}
\end{figure*}

To test how large a data size is required to estimate a class-based
model, we conducted the following experiment. We defined an artificial
class-based model and generated some data according to its
distribution. We then used the data to estimate a class-based model
(dendroid distribution), and evaluated the estimated model by
measuring the number of dependencies (dependency arcs) it has and the
KL distance\footnote{The KL distance (KL divergence or relative
  entropy) which is widely used in information theory and statistics,
  is a measure of `distance' between two distributions \cite{Cover91}.
  It is always non-negative and is zero if and only if the two
  distributions are identical, but is asymmetric and hence not a
  metric (the usual notion of distance).} between the estimated model
and the true model. We repeatedly generated data and observed the
`learning curve,' namely the relationship between the number of
dependencies in the estimated model and the data size used in
estimation, and the relationship between the KL distance between the
estimated and true model and the data size. We defined two other
models and conducted the same experiments. Figure \ref{fig:sim} shows
the results of these experiments for these three artificial models
averaged over $10$ trials. (The number of parameters in Model1,
Model2, and Model3 are $18$, $30$, and $44$ respectively, while the
number of dependencies are $1$, $3$, and $5$ respectively.) We see
that to accurately estimate a model the data size required is as large
as $100$ times the number of parameters. Since a class-based model
tends to have more than $100$ parameters usually, the current data
size available in the Penn Tree Bank (See Table\ref{tb:verb}) is not
enough for accurate estimation of the dependencies within case frames
of most verbs.

\section{Conclusions}

We conclude this paper with the following remarks.
\begin{enumerate}
\item The primary contribution of research reported in this paper is
  that we have proposed a method of learning dependencies between case
  frame slots, which is theoretically sound and efficient, thus
  providing an effective tool for acquiring case dependency
  information. 
\item For the slot-based model, sometimes case slots are found to be
  dependent. Experimental results demonstrate that using the
  dependency information, when dependency does exist, structural 
  disambiguation results can be improved.
\item For the word-based or class-based models, case slots are judged
  independent, with the data size currently available in the Penn Tree
  Bank. This empirical finding verifies the independence assumption
  widely made in practice in statistical natural language processing.
\end{enumerate}

We proposed to use dependency forests to represent case frame
patterns. It is possible that more complicated probabilistic
dependency graphs like Bayesian networks would be more appropriate for
representing case frame patterns. This would require even more data
and thus the problem of how to collect sufficient data would be a
crucial issue, in addition to the methodology of learning case frame
patterns as probabilistic dependency graphs. Finally the problem of
how to determine obligatory/optional cases based on dependencies
acquired from data should also be addressed.

\section*{Acknowledgement}

We thank Mr.K.Nakamura, Mr.T.Fujita, and Dr.K.Kobayashi of NEC C\&C
Res. Labs. for their constant encouragement. We thank Mr.R.Isotani of
NEC Information Technology Res. Labs. for his comments. We thank Ms. 
Y.Yamaguchi of NIS for her programming effort.

\end{document}